\documentclass[letterpaper, 10 pt, conference]{ieeeconf}  

\IEEEoverridecommandlockouts                              

\usepackage{graphics} 
\usepackage{epsfig} 
\usepackage{mathptmx} 
\usepackage{times} 
\usepackage{amsmath} 
\usepackage{amssymb}  
\usepackage{booktabs}
\usepackage{multirow}
\usepackage[font=small,skip=0pt]{caption}

\title{\LARGE \bf
GO-FEAP: Global Optimal UAV Planner Using Frontier-Omission-Aware Exploration and Altitude-Stratified Planning
}

\author{\&Weiye Zhang$^{1}$,\&Wenshuai Yu$^{1}$, Licong Zhuang$^{2}$, Xiaoyi Zhang$^{2}$, Zhi Zeng$^{3}$ and Jiasong Zhu$^{1}$$^{*}$
\thanks{*This work was not supported by any organization}
\thanks{\&These authors contributed equally to this work and should be considered co-first authors.}
\thanks{$^{1}$Weiye Zhang, Wenshuai Yu, Jiasong Zhu are with College of Civil and Transportation Engineering, Shenzhen University, Shenzhen 518060, China.
 {\tt\small zhangweiye2021@email.szu.edu.cn, ywsh@szu.edu.cn, zjsong@szu.edu.cn}}%
\thanks{$^{2}$Licong Zhuang and Xiaoyi Zhang are with Guangdong Laboratory of Artificial Intelligence and Digital Economy (SZ), Shenzhen 518060, China.
        {\tt\small zhuanglicong@gml.ac.cn, zhangxiaoyi@gml.ac.cn}}%
\thanks{$^{3}$Zhi Zeng is with College of Architecture and Urban Planning, Shenzhen University, Shenzhen 518060, China.
        {\tt\small cengzhi2021@email.szu.edu.cn}}%
}

\begin{document}
\captionsetup{font=small,skip=0pt} 

\maketitle
\thispagestyle{empty}
\pagestyle{empty}

\begin{abstract}

Autonomous exploration is a fundamental problem for various applications of unmanned aerial vehicles(UAVs). Existing methods, however, are demonstrated to static local optima and two-dimensional exploration. 
To address these challenges, this paper introduces GO-FEAP (Global Optimal UAV Planner Using Frontier-Omission-Aware Exploration and Altitude-Stratified Planning), aiming to achieve efficient and complete three-dimensional exploration. 
Frontier-Omission-Aware Exploration module presented in this work takes into account multiple pivotal factors, encompassing frontier distance, nearby frontier count, frontier duration, and frontier categorization, for a comprehensive assessment of frontier importance. 
Furthermore, to tackle scenarios with substantial vertical variations, we introduce the Altitude-Stratified Planning strategy, which stratifies the three-dimensional space based on altitude, conducting global-local planning for each stratum. 
The objective of global planning is to identify the most optimal frontier for exploration, followed by viewpoint selection and local path optimization based on frontier type, ultimately generating dynamically feasible three-dimensional spatial exploration trajectories.
We present extensive benchmark and real-world tests, in which our method completes the exploration tasks with unprecedented completeness compared to state-of-the-art approaches.

\end{abstract}

\section{INTRODUCTION}

Unmanned Aerial Vehicles (UAVs), especially quadrotors have gained widespread popularity within the realm of robotics, demonstrating immense potential across a spectrum of applications, including industrial inspections \cite{nooralishahi2021drone} \cite{zhang2018autonomous}, search and rescue \cite{yang2020maritime} \cite{ribeiro2021unmanned}. Among these tasks in complex environment, such as autonomous exploration of unknown areas, have garnered extensive attention in the context of UAV applications.

The critical challenge in these scenarios lies in devising exploration planning methods that not only consider environmental intricacies but also harness the UAV's maneuverability and autonomous capabilities to achieve globally optimal paths. In recent years, various exploration planning methods \cite{cieslewski2017rapid} \cite{schmid2020efficient} \cite{meng2017two} \cite{zhou2021fuel} \cite{wittinghistory} \cite{Exploration-RRT} have been proposed with some real-world experiments presented. However, existing methods have exhibited two noteworthy shortcomings: static local optima and two-dimensional exploration. Firstly, many approaches rely on static local optima, optimizing for frontier information at the time of each local path planning instance. While frontier are subject to real-time fluctuations, rendering the utilization of static optima often prone to overlooking exploration at the global level. Secondly, some methods primarily concentrate on two-dimensional exploration, neglecting the fact that certain unknown spaces exhibit substantial variations in the vertical dimension. These methodologies may prove inadequate when confronted with practical scenarios involving substantial vertical span.

In order to address the aforementioned challenges, this paper proposes \textbf{GO-FEAP}, \textbf{G}lobal \textbf{O}ptimal UAV Planner Using \textbf{F}rontier-Omission-Aware \textbf{E}xploration and \textbf{A}ltitude-Stratified \textbf{P}lanning, aiming at achieving efficient three-dimensional spatial exploration. During the UAV's flight exploration, where environment information and UAV's positions are continually updated, the \textit{Frontier-Omission-Aware Exploration module} presented in this work takes into account multiple pivotal factors related to frontier. These factors include frontier distance, the number of neighboring frontiers, frontier duration, and frontier classification, all of which are comprehensively assessed to determine the frontier importance. Furthermore, to tackle scenarios with substantial vertical variations, we introduce the \textit{Altitude-Stratified Planning strategy}, which stratifies the three-dimensional space based on altitude, conducting global-local planning for each altitude stratum. The objective of global planning is to identify the most optimal frontier for exploration, followed by viewpoint selection and local path optimization based on frontier type. This approach ultimately results in the generation of dynamically feasible three-dimensional spatial exploration trajectories.

The primary contributions of this research can be summarized as follows:

1) A Frontier-Omission-Aware Exploration module 
evaluates and prioritizes frontiers, accounting for key factors and enhancing UAV exploration planning adaptability to real-time frontiers changes.

2) An Altitude-Stratified Planning strategy simultaneously  addresses scenarios with substantial vertical variations by partitioning the three-dimensional space by altitude, resulting in more effective path planning in complex environment.

3) Extensive simulation and real-word tests demonstrate the effectiveness of the proposed GO-FEAP methods. 


\section{Related Work}
In the realm of autonomous exploration, a number of prior research efforts can be broadly categorized into two major classes: sampling-based exploration and frontier-based exploration.

Sampling-based exploration, a major category among proposed methods, generates viewpoints randomly to explore space, closely related to the concept of the next best view (NBV) \cite{connolly1985determination}, which repeatedly computes covering views to construct a complete scene model. \cite{bircher2016receding} first applied NBV in 3D exploration, expanding rapidly-exploring random trees (RRTs) within accessible space and executing the edge with the highest information gain in a receding horizon fashion. This method was extended to consider factors such as localization uncertainty \cite{papachristos2017uncertainty}, visual importance of objects \cite{dang2018visual}, and inspection tasks \cite{bircher2018receding}. To prevent discarding expanded trees, roadmaps were constructed in \cite{wittinghistory} \cite{wang2019efficient} to reuse prior knowledge, while \cite{schmid2020efficient} continuously maintains and refines a single tree using a rewiring scheme inspired by RRT*. Additionally, \cite{dharmadhikari2020motion} samples safe and dynamically feasible motion primitives directly and executes the most informative one to achieve faster flight.

Frontier-based approaches, another classic category, were first introduced in \cite{yamauchi1997frontier} and later subjected to comprehensive evaluation in \cite{julia2012comparison}. In the original method \cite{yamauchi1997frontier}, the closest frontier is selected as the next target. However, \cite{cieslewski2017rapid} introduced a novel approach that selects the frontier within the field of view (FOV) in each decision, minimizing velocity changes to maintain a consistently high flight speed. This scheme outperforms the classic method \cite{yamauchi1997frontier}. Furthermore, \cite{deng2020robotic} introduced a differentiable measure of information gain based on frontiers, enabling path optimization with gradient information.

Certain approaches combine the strengths of frontier-based and sampling-based methods. For instance, \cite{selin2019efficient} devise global paths toward frontiers and locally sample paths. Furthermore, \cite{charrow2015information} presents a gradient-based approach to optimize the local path. \cite{meng2017two} samples viewpoints around frontiers and computes the global shortest tour passing through them, while \cite{song2017online} employs a sampling-based algorithm to generate inspection paths that comprehensively cover the frontiers.

However, most existing methods tend to optimize for the frontier at each time of planning without considering the changing frontier environment
, resulting in omission. In contrast, our approach consider multiple critical factors to enhance the adaptability of UAV exploration to change real-time frontiers.

What's more, 
although some approaches\cite{selin2019efficient}\cite{cieslewski2017rapid} attempt to shift their attention towards three-dimensional environment, they often merely apply two-dimensional methods mechanically to the three-dimensional setting, without adequately considering the unique characteristics of three-dimensional scenes. 
The significant variations in altitude not only presents challenges for flight control, but can also lead to a decrease in task execution efficiency.


\section{Proposed Method}
\begin{figure}
\begin{center}

\resizebox*{1.0\linewidth}{!}{\includegraphics{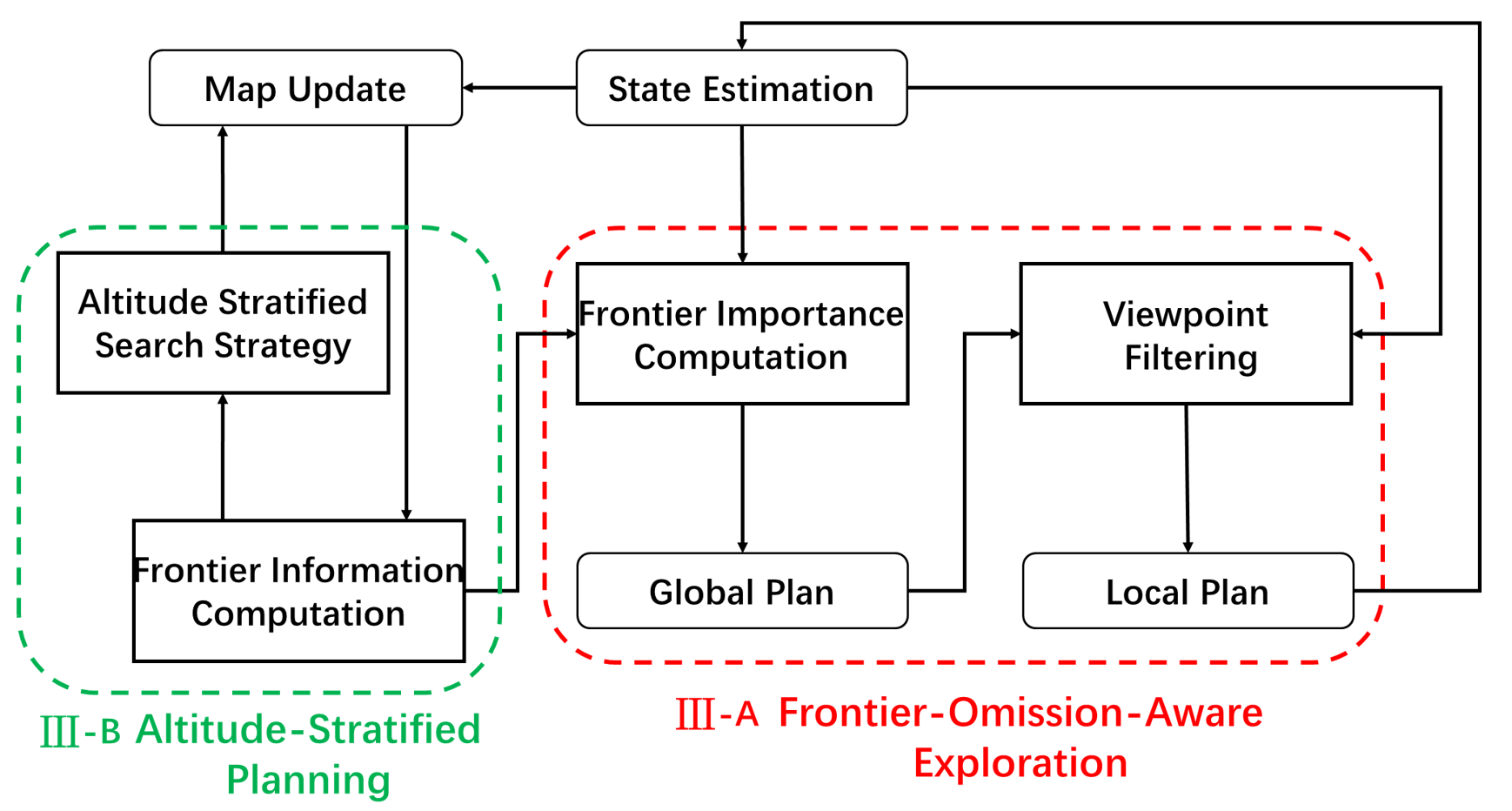}}\hspace{5pt}
\setlength{\abovecaptionskip}{-0.2cm}
\setlength{\belowcaptionskip}{-0.6cm}
\caption{An overview of the proposed GO-FEAP framework.}
\label{overview}
\end{center}
\end{figure}

In this research, we propose GO-FEAP, a global optimal planner leveraging frontier-omission-aware exploration and altitude-stratified planning, to enhance the efficiency of UAV autonomous exploration in unknown areas. An overview of the proposed approach is presented in Figure \ref{overview}.

Map is updated as the UAV moves. If the map update alters frontiers, new frontiers are extracted, and the computation of frontier information follows. (Section \ref{4.1}). Subsequently, a utility function is created to assess frontier importance and directs global planning(Section \ref{4.2}). After that, the generation of reasonable viewpoints and paths is optimized to guide continuous UAV exploration (Section \ref{4.3}).
The process continues until the designated space within a specified height is fully explored. If there are still unknown areas at higher altitudes, the exploration plane's height is raised until all unknown spaces are explored (Section \ref{4.4}).

\subsection{Frontier-Omission-Aware Exploration Module}
\subsubsection{Frontier Information Computation and Update}
\label{4.1}
In GO-FEAP, \cite{han2019fiesta} is employed as the occupancy mapping server, which is suitable for integrating sensor data to represent free space, and the maintained ESDF structure allows for easy planning of collision-free paths.

 Assuming the center of frontier A is $\left(X_A, Y_A\right)$, and the center of frontier B is $\left(X_B, Y_B\right)$, with the maximum sampling distance of $r$ and the sampling interval of $\Delta \theta$. The probability that a B sampling point is closer to the center of frontier A is denoted as $P_{\text {error }}$, which is evaluated as equation \ref{1}. When the distance $d_{a, b}$ between the centers of frontier A and B is less than the maximum sampling distance $r$, there is a high probability that a sampling point from frontier A is closer to the center of frontier B.

\begin{equation}
\setlength{\abovedisplayskip}{-0.7cm}
P_{\text {error }}= \begin{cases}\frac{\theta r-r \cdot \sin \theta}{4 \cdot \sin (\theta / 2)}, & d_{a, b}<r \\ 0, & d_{a, b}>r\end{cases} \label{1}
\end{equation}

\begin{equation}
d_{a, b}=\sqrt{\left(X_a-X_b\right)^2+\left(Y_a-Y_b\right)^2}
\end{equation}

\begin{figure}
\begin{center}
\resizebox*{1.0\linewidth}{!}{\includegraphics{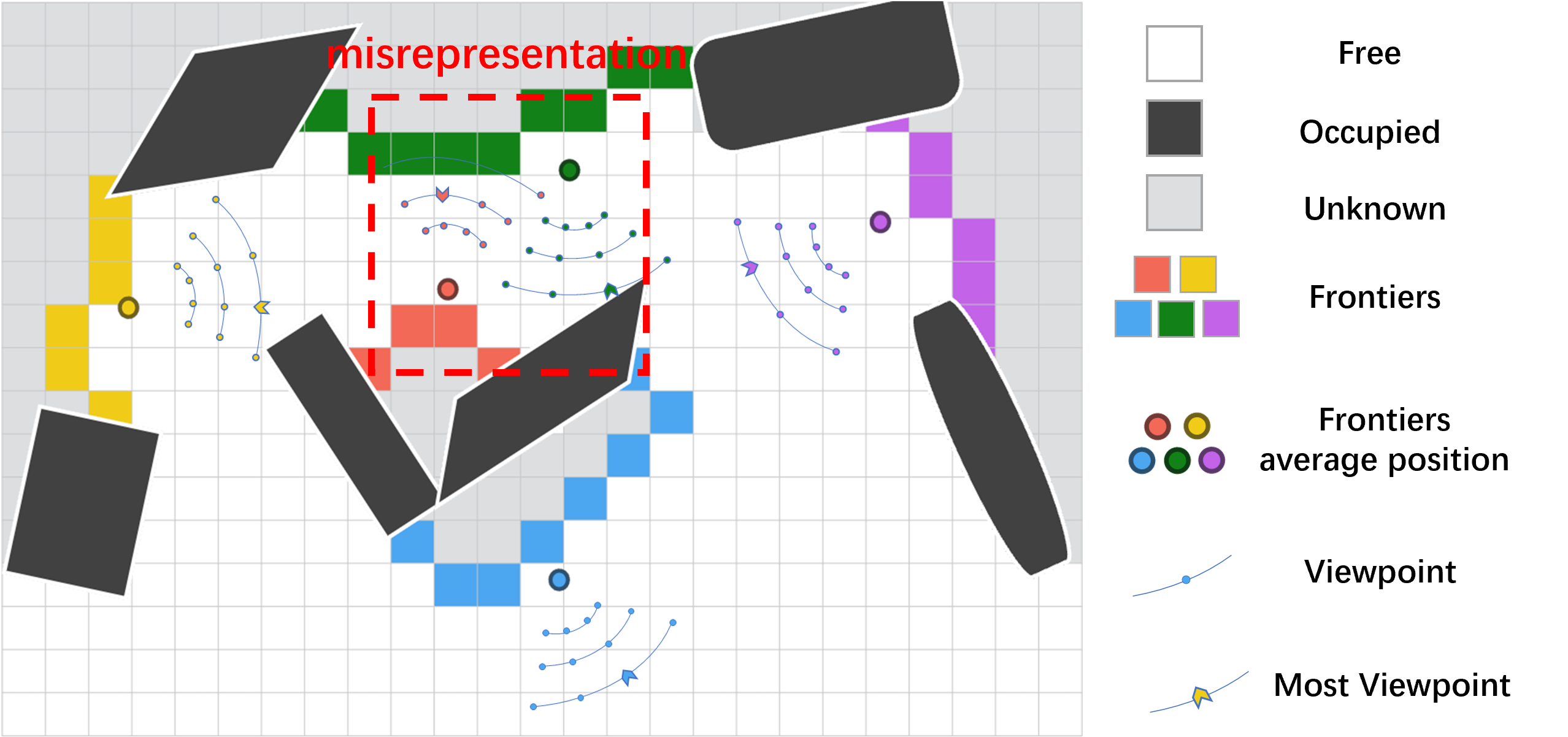}}\hspace{5pt}
\setlength{\belowcaptionskip}{-0.5cm}
\caption{Comparing frontier viewpoint: most viewpoint vs. average position. FUEL \cite{zhou2021fuel} uses viewpoints to represent frontier, resulting in misrepresentation, while GO-FEAP avoid it utilizing the average position.}
\label{2}
\end{center}
\end{figure}

As shown in Figure \ref{2}, the average position of frontiers can more accurately reflect the mutual relationship between frontiers, compared to calculating frontier distances using viewpoints.

The A* algorithm is employed for calculating the distance between frontiers, determining the number $N_{\text{near}, i}$ of nearby frontiers, and recording the duration $T_i$ of each frontier. This facilitates the assessment of the possibility $\xi_{f_i}$ of frontier being missed. The data contained within the frontier data structure is presented in Table \ref{Frontier}.

\begin{table}
\centering
\renewcommand{\arraystretch}{1.1}

\caption{Frontier data structure.}
\label{Frontier}
\begin{tabular}{ll}
\toprule
\textbf{Data}                                          & \textbf{Explanation} \\
\hline
$C_i$          & Frontier cells     \\
$P_{\text {avg }, i}$      & Average position of $C_i$   \\
$B_i$   & Axis-aligned bounding box of $C_i$  \\
$N_{\text {near }, i}$  & Number of nearby frontier cells  \\
$T_i$  & Duration time of $C_i$ \\
$I_{\text {cost }, i}$   & Importance cost between $C_i$ and other cells   \\
$V P_i$ & Effective candidate viewpoints of $C_i$\\
\bottomrule
\end{tabular}
\end{table}

\subsubsection{Global Planning Considering Frontier Omission}
\label{4.2}

To find the optimal exploration order for frontiers, the proposed GO-FEAP takes inspiration from the Traveling Salesman Problem \cite{poikonen2019branch}, which aims to find the shortest path connecting all targets. By designing a suitable cost matrix, existing algorithmic solutions can be applied for solving the problem. 
The important assessment of each frontier is achieved by incorporating the distance between frontiers, the size of frontier cells, the count of nearby frontiers, and frontier duration into a metric. Subsequently, this allows the construction of a cost matrix. Assuming there exist $N_f$ frontiers at any given moment, the dimensions of the cost matrix are $(N_f+1)\times(N_f+1)$. In order to transform it into a symmetric Traveling Salesman Problem, the first column of the cost matrix is set to \textbf{0}, while the remaining elements are determined as per equation \ref{matrix}.

\begin{equation}
\setlength{\abovedisplayskip}{-0.2cm}
\setlength{\belowdisplayskip}{0.1cm}
M_{t s p}\left(x_i, x_j\right)= \begin{cases}I_{\text {cost }}^p\left(C_i, C_j\right) & i=0, j \in\left\{1,2, \ldots, N_f\right\} \\ I_{\text {cost }}^f\left(C_i, C_j\right) & i \neq 0, j \in\left\{1,2, \ldots, N_f\right\}\end{cases}\label{matrix}
\end{equation}

\begin{figure}
\begin{center}
\resizebox*{1.0\linewidth}{!}{\includegraphics{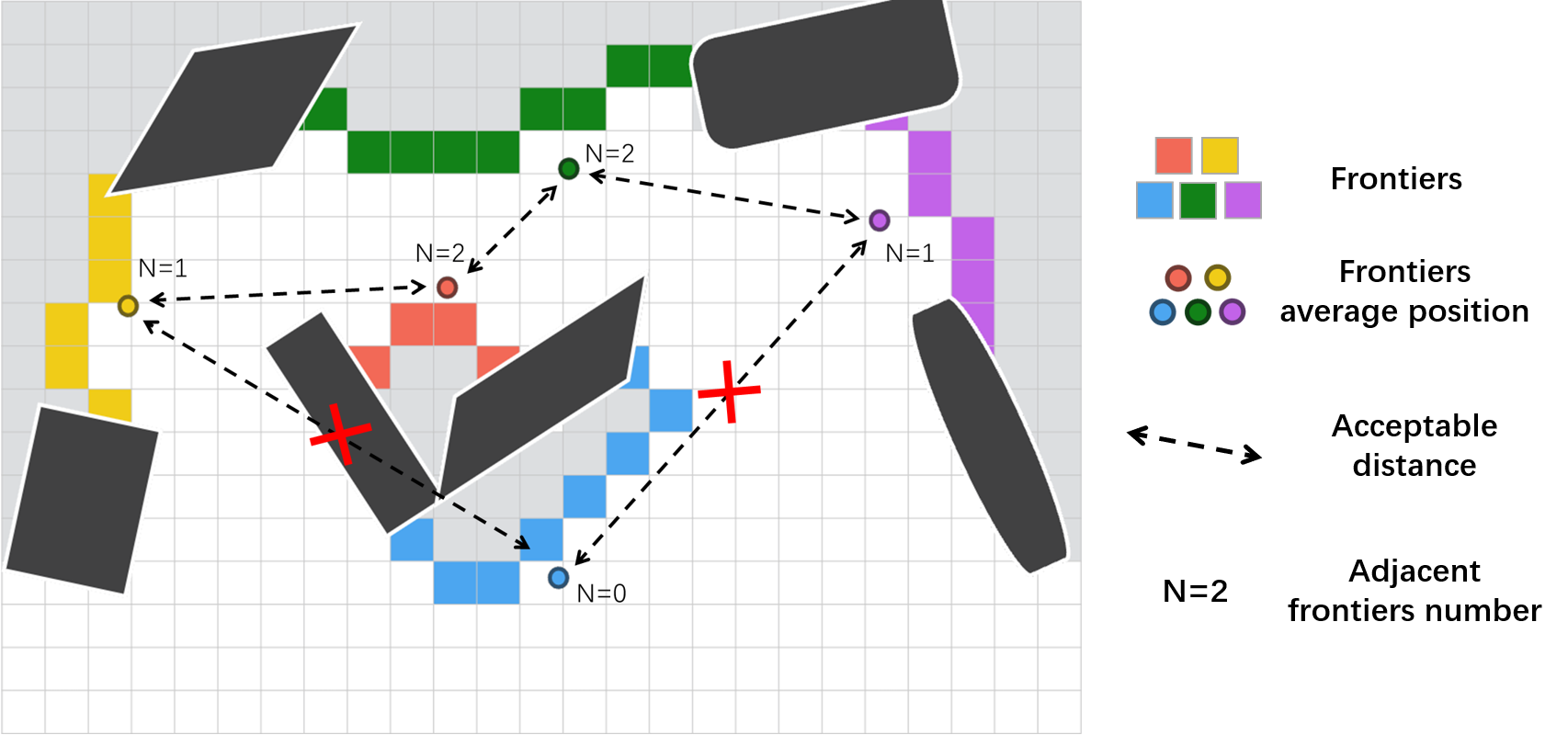}}\hspace{5pt}
\setlength{\belowcaptionskip}{-1.0cm}
\caption{Illustration of the calculation method for nearby frontiers counts}

\label{3}
\end{center}
\end{figure}

The idea of dynamic programming is taken by us and applied to the exploration problem, as demonstrated in equation \ref{basic}. In the basic paradigm \ref{basic}, we incorporate considerations for the possibility of frontier omissions, as seen in equation \ref{basic2}.
\begin{equation}
\setlength{\abovedisplayskip}{0.2cm}
\setlength{\belowdisplayskip}{-0.1cm}
d p[i][a]=\max \left(d p[i-1]\left[A-A_{f_k}\right]+C_{f_k}, \ldots\right) \label{basic}
\end{equation}

\begin{equation}
\setlength{\abovedisplayskip}{0cm}
d p[i][a]=\max \left(d p[i-1]\left[A-A_{f_k}\right]+C_{f_k}+\xi_{f_k}, \ldots\right) \label{basic2}
\end{equation}

The importance level of frontier $I_{\text {cost }}^f\left(C_i, C_j\right)$ is computed by the combination of the distance between frontiers $S\left(P_{\text {avg }, i}, P_{\text {avg }, j}\right)$ and the possibility of frontier omissions $\xi_f$ using equation \ref{cost}. When calculating the current position $P_\text {uav }$, the equation is as follows \ref{cost2}:

\begin{equation}
\setlength{\abovedisplayskip}{-0.1cm}
\begin{aligned}
& I_{\text {cost }}^f\left(C_i, C_j\right)=\operatorname{Cost}_f+\xi_f \\
& =S\left(P_{\text {avg }, i}, P_{\text {avg }, j}\right) / V_{\max }+\alpha_f \cdot \omega_f\left(T_i\right) \cdot \beta_f\left(N_{\text {near }, i}\right)
\label{cost}
\end{aligned}
\end{equation}

\begin{equation}
\begin{aligned}
& I_{\text {cost }}^p\left(C_i, C_j\right)=\operatorname{Cost}_pf+\xi_pf \\
& =S\left(P_{\text {uav }}, P_{\text {avg }, j}\right) / V_{\max }+\alpha_p \cdot \omega_p\left(T_i\right) \cdot \beta_p\left(N_{\text {near }, i}\right)
\label{cost2}
\end{aligned}
\end{equation}


During the calculation process, the frontier that satisfy 
$S\left(P_{\text {uav }}, P_{\text {avg }, j}\right)<S_{\text {near }}$ are collected and added to the list NearCur\_IDs (as shown in Figure \ref{3}). 
$\alpha_p$ is related to the number of cells associated with the frontier. 
$\omega_p$ imposes constraints on the duration of the frontier. 
Meanwhile, $\beta_f$ and $\phi$ respectively limit the number and distance of frontiers near the frontier as shown equation \ref{obp}.
\begin{equation}
\begin{gathered}
\omega_p\left(T_i\right)= \begin{cases}\omega_i, & 0<T_i<t_{\max } \\
0, & t_{\max } \leq T_i\end{cases} \\
\beta_f\left(N_{\text {near }, i}\right)= \begin{cases}N_t, & 0<N<n_{\max } \\
0, & n_{\max } \leq N_{\text {near }, i}\end{cases} \\
\phi(S)= \begin{cases}1, & 0<S<S_{\text {near }} \\
0, & S \geq S_{\text {near }}\end{cases}
\label{obp}
\end{gathered}
\end{equation}

The importance assessment takes into account not only the distance between frontiers but also incorporates predictive factors for future exploration. This allows the generated path to consider not only the current frontier positions but also the potential elimination of local omissions, avoiding revisiting previously explored areas in final stages.

\begin{figure}
\begin{center}
\resizebox*{1.0\linewidth}{!}{\includegraphics{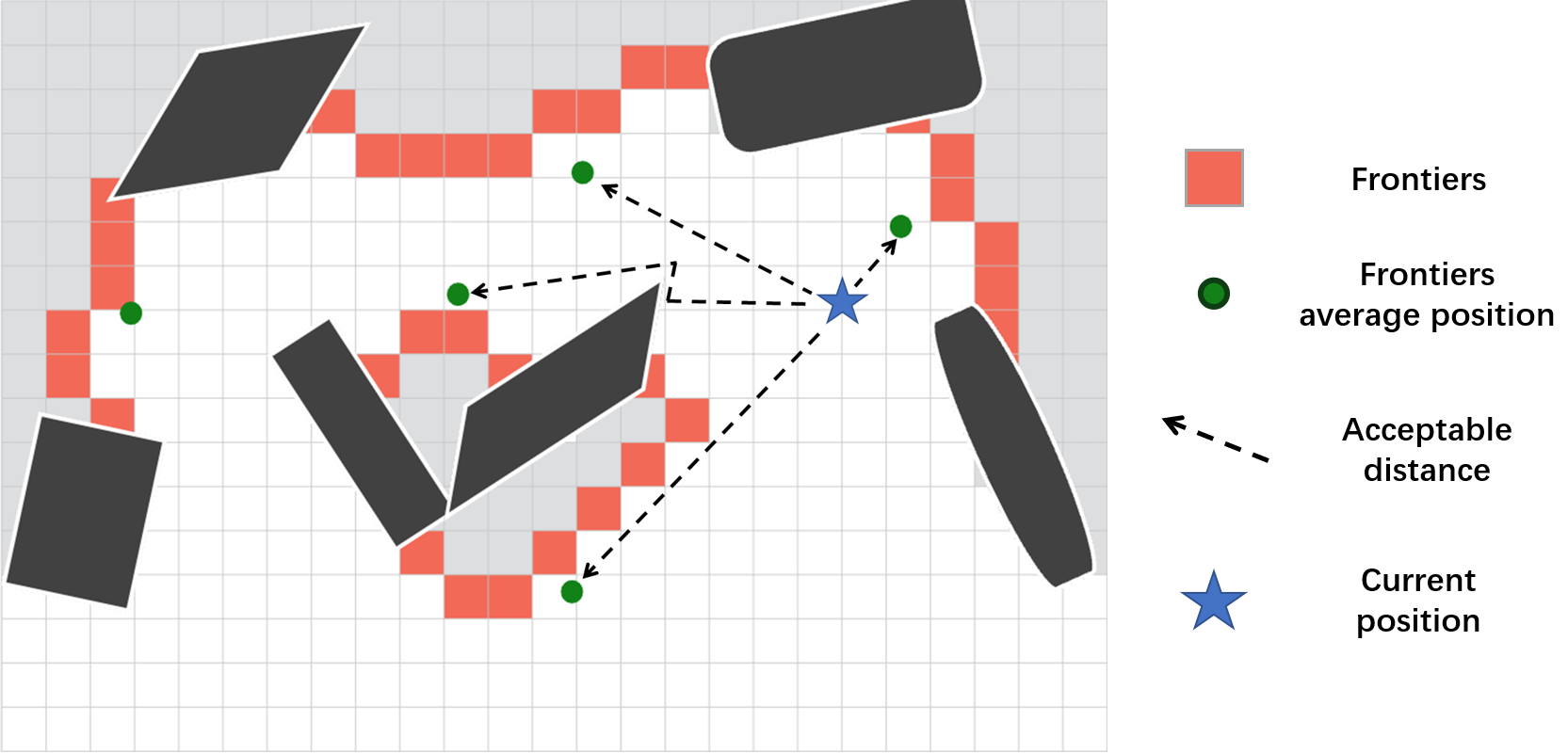}}\hspace{5pt}
\setlength{\abovecaptionskip}{-0.2cm}
\setlength{\belowcaptionskip}{-0.6cm}
\caption{Global path planning based on local frontiers.}

\label{4}
\end{center}
\end{figure}

Due to the continuous changes in frontier, a departure from the Traveling Salesman Problem (TSP) \cite{poikonen2019branch}, which necessitates the computation of a global path connecting all frontiers, is undertaken. Instead, the consideration is limited to x frontiers situated around the current pose of the UAV, as depicted in Figure \ref{4}. This not only extends the applicability of our method to larger scenarios but also mitigates exploration inconsistencies stemming from the need for open or closed-loop tours inherent to the Traveling Salesman Problem. This approach ensures a higher probability of continuity in each exploration instance amidst fluctuations in frontier information. We utilize the previously calculated NearCur\_IDs to compress the cost matrix constructed earlier. It is important to note that when the number of frontier satisfying $S\left(P_{\text {uav }}, P_{\text {avg }, j}\right)<S_{\text {near }}$ is less than the minimum quantity $N_{\min }^{t s p}$ required to the symmetric Traveling Salesman Problem, the closest frontier will be selected for filling, even if it is far from the current position. The specific formula for matrix compression is as follows. The specific formula for matrix compression is as equations \ref{M}.

\begin{equation}
\setlength{\abovedisplayskip}{-0.1cm}
\setlength{\belowdisplayskip}{0.2cm}
M_{t s p}^{c o m p}\left(x_{i d, i}, x_{i d, j}\right)=M_{t s p}\left(x_i, x_j\right), \quad i, j \in NearCur\_IDs \label{M}
\end{equation}

\subsubsection{Local Planning Based on Frontier Region}
\label{4.3}
After calculating the global sequence of a certain number of neighboring frontiers for observation using the variant TSP, it is necessary to further determine the optimal viewpoint for each frontier. To achieve this, we draw inspiration from the graph search methods in \cite{mahapatra1997scalable} and apply a filtering process to the candidate viewpoint set based on the characteristics of each frontier. By considering the order of the frontiers to be observed and the information derived from the filtered viewpoint set, we optimize the generation of an optimal observation path. This path guides the exploration of the highest-priority exploration frontier. Due to certain special frontier conditions, using a unified observable quantity as the evaluation criterion for viewpoint filtering may lead to unnecessary path length and travel time. Therefore, we consider adopting different viewpoint filtering methods based on the properties of the highest-priority exploration frontier.

A sliding window is maintained to compute the average percentage of obstacles within a specific time range, approximating potential scenarios for the highest-priority exploration frontier. In each frame, an evaluation is performed on the point cloud generated from the input depth map. Points within a distance less than the maximum perception range from the camera center are classified as obstacles, while those beyond this range are categorized as free space. The proportion $\operatorname{flag}_i$ of obstacles in each frame is calculated, and the sliding window \textit{flag\_que} is updated accordingly.

\begin{figure}
\begin{center}
\resizebox*{1.0\linewidth}{!}{\includegraphics{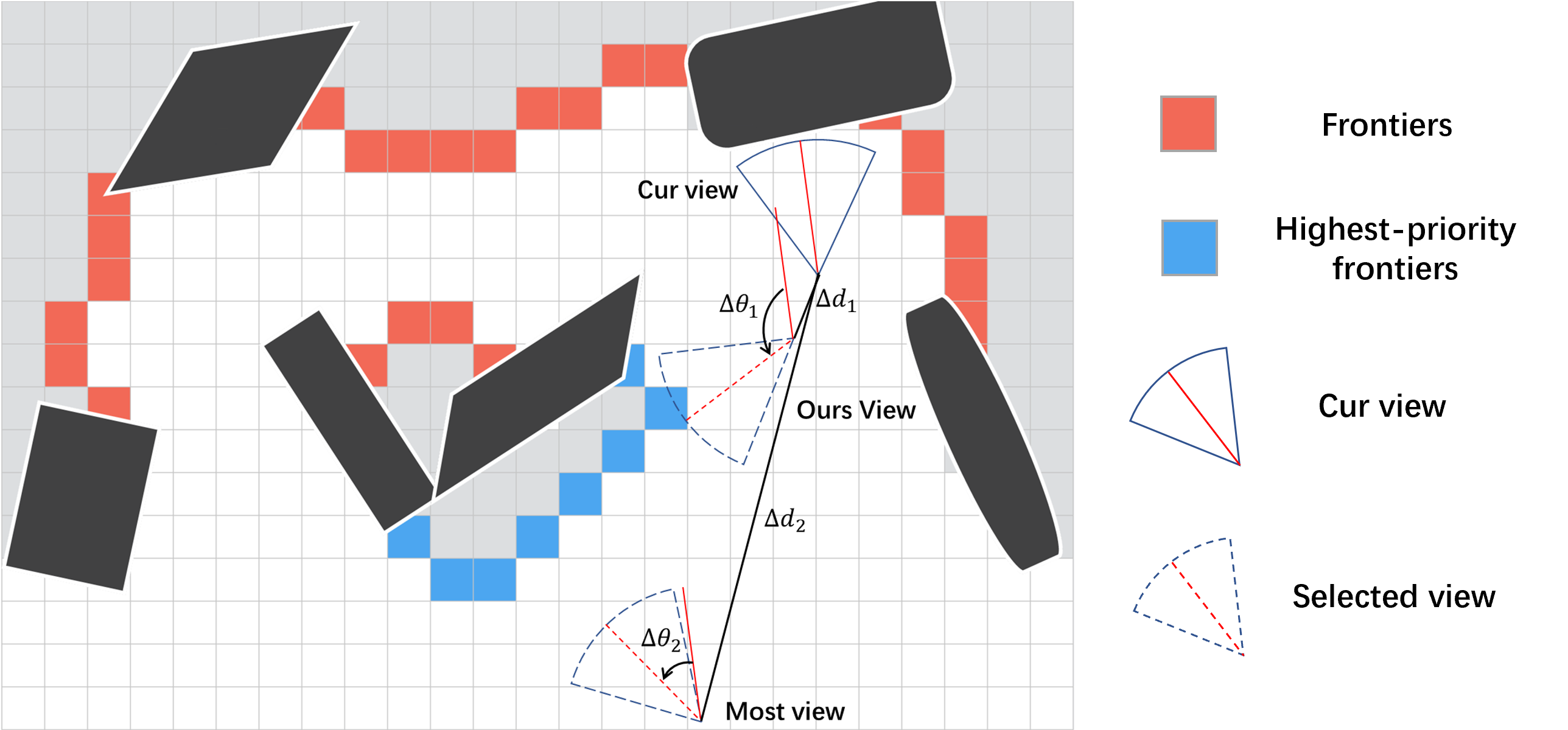}}\hspace{5pt}
\setlength{\abovecaptionskip}{-0.2cm}
\setlength{\belowcaptionskip}{-0.6cm}
\caption{Local optimization with high deviation viewpoints. prioritizing high deviation viewpoints for nearby target frontier.}

\label{6}
\end{center}
\end{figure}

As depicted in Figure \ref{6}, when the highest-priority exploration frontier is relatively close to the current position, a preference for steering rather than a simple repositioning to explore the frontier is observed as more effective. Consequently, candidate viewpoints that are both in closer proximity to the current position and exhibit a greater steering angle are selected. In regions characterized by corners, a preference for generating more oscillatory head turns is established to facilitate the efficient exploration of narrow spaces. In light of these scenarios, the evaluation criterion for filtering is considered to be the expected steering rate $\varphi_{V P_i}=\operatorname{cost}_{\text {yaw }} / \operatorname{cost}_{\text {dist }}$ of the viewpoint. In cases where none of the aforementioned conditions are met, filtering based on the observable quantity $N_{\text {view }}$ is employed. 

A directed acyclic graph is created by forming graph nodes based on the filtered set of viewpoints. The optimal local path is searched for using the Dijkstra algorithm, and the optimal viewpoint for the highest-priority exploration frontier is obtained. Then, a smooth, safe, and dynamically feasible B-spline trajectory is generated using the method described in \cite{zhou2019robust}. All parameters of the B-spline curve are further optimized to minimize the total trajectory time, allowing the quadcopter to fully utilize its dynamic capabilities.

\subsection{Altitude-Stratified Planing Strategy}
\label{4.4}
\begin{figure}
\begin{center}
\resizebox*{0.6\linewidth}{!}{\includegraphics{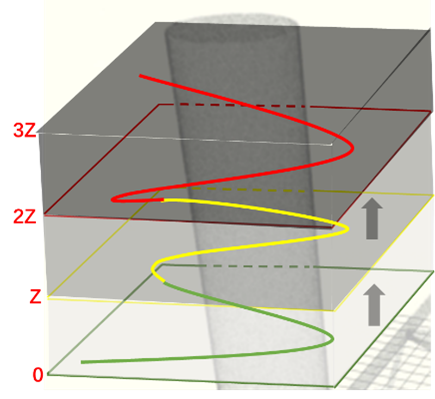}}\hspace{5pt}
\setlength{\abovecaptionskip}{0.1cm}
\setlength{\belowcaptionskip}{-0.4cm}
\caption{Altitude-Stratified Search Strategy.Increase the exploration height after finishing each layer to explore the entire space.}

\label{8}
\end{center}
\end{figure}

To enable exploration in three-dimensional space, we implement an altitude stratified planing strategy (Figure \ref{8}). Initially, a limited exploration height is set to ensure that there are no targets at different heights within the same horizontal position, allowing more efficient exploration speed. Once the exploration of a particular layer is completed, the exploration height is uniformly increased in both upward and downward directions. For example, if the exploration height is set to "$z$" the search range for the first layer is $[0, z]$. Upon completing the exploration of the first layer, the exploration height is updated to $[z, 2z]$. This process is repeated, continuously increasing the exploration height and completing the exploration of each layer until the predetermined maximum exploration height is reached. Please refer to the accompanying diagram for a visual representation.

When a height range is specified, the UAV operates at different attitude due to considerations of its dynamics. Additionally, the upper boundary on the depth map perceived by the UAV exceeds the height limit of the current layer. 
So, when the exploration plane are elevated, we retain information from the prior height level, using it as prior knowledge to guide exploration at the current level. This efficient use of data from previous heights improves exploration efficiency.

\section{Simulation Results}
To comprehensively evaluate the proposed GO-FEAP method, we test our proposed framework in simulation. We assess it in a powerplant scenario and a maze scenario for their different scale.

\subsection{Implementation Details}
We utilize a simulation environment built upon the map structure based on \cite{han2019fiesta} and the UAV model from \cite{lee2010geometric}, which enables the integration of perception, mapping, estimation, and control into a closed-loop system. The proposed GO-FEAP method is compared with classical algorithms presented in \cite{cieslewski2017rapid} \cite{zhou2021fuel} \cite{selin2019efficient}. All methods are run five times using the same configuration.

All experiments are conducted with uniform settings: the resolution of the occupancy map is $r=0.1m$, the maximum sensing range of depth sensors $d_{\max }^{\text {sensor }}$ is set as $4.5m$, the FOVs of vertical and horizontal direction are set as $[80 \times 60]$ deg. And we also set the dynamic limitations including the maximum flight speed of the UAV is$\quad v_{\max }=2 \mathrm{~m} / \mathrm{s}$, the maximum angular velocity is$\quad \dot{\xi}_{\max }=1 \mathrm{rad} / \mathrm{s}$.

\subsection{Maze Environment Simulation}

The first maze simulation scenario is $20 \times 10 \times 3 \mathrm{~m}^3$ large, a closed and bounded low-height space, which is suitable for testing our proposed Frontier-Omission-Aware Exploration planning module.

\begin{table}[]
\caption{Exploration data statistics in maze environment}
\label{tabmaz}
\renewcommand{\arraystretch}{1.2}
\begin{tabular}{ccccccccc}
\hline
\multirow{2}{*}{\textbf{Method}} & \multicolumn{4}{c}{\textbf{Exploration time}}             & \multicolumn{4}{c}{\textbf{Flight distance}}              \\ \cline{2-9} 
                                 & \textbf{Avg} & \textbf{Std} & \textbf{Max} & \textbf{Min} & \textbf{Avg} & \textbf{Std} & \textbf{Max} & \textbf{Min} \\ \hline
Rapid                     & 499          & 42           & 570          & 460          & 276          & 16           & 298          & 257          \\ 
AEP                       & 342          & 37           & 392          & 295          & 155          & 17           & 182          & 136          \\ 
FUEL                      & 141          & 8            & 152          & 132          & 192          & 5            & 200          & 188          \\ 
Ours                       & 125          & 9            & 133          & 114          & 146          & 13           & 160          & 130          \\ \hline
\end{tabular}
\end{table}

\begin{figure}
\begin{center}
\resizebox*{1.0\linewidth}{!}{\includegraphics{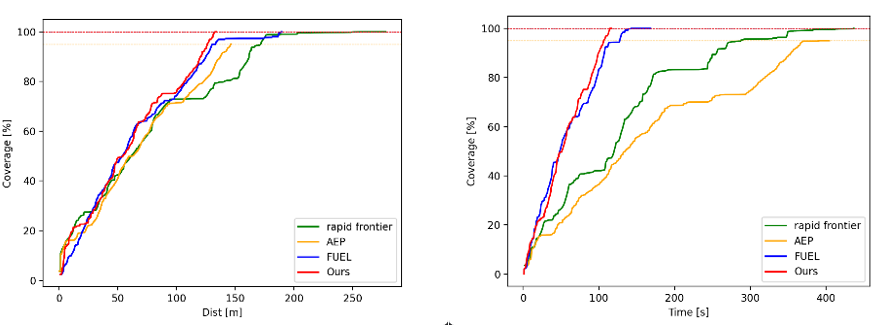}}\hspace{5pt}
\setlength{\abovecaptionskip}{-0.3cm}
\setlength{\belowcaptionskip}{-0.6cm}
\caption{Comparing exploration progress in maze environment.}

\label{9}
\end{center}
\end{figure}

\begin{figure}
\begin{center}
\resizebox*{1.0\linewidth}{!}{\includegraphics{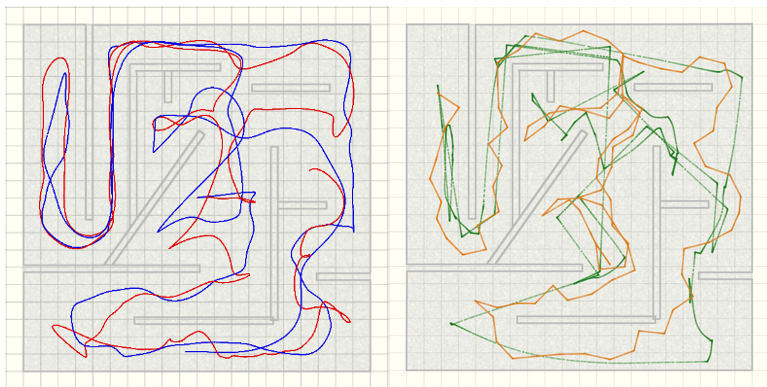}}\hspace{5pt}
\setlength{\abovecaptionskip}{-0.2cm}
\setlength{\belowcaptionskip}{-0.6cm}
\caption{Trajectories of GO-FEAP, FUEL, AEP, and rapid frontier in maze environment, shown as red, blue, green and orange respectively.}

\label{10}
\end{center}
\end{figure}

The statistical results of the exploration time and distance are shown in the Table \ref{tabmaz}. The exploration progress over time is depicted in Figure \ref{9}, and the exploration trajectories are illustrated in the Figure \ref{10}.

The following observations can be made from Figure \ref{9}: 1)During the early stage of exploration, all four methods exhibit the fastest increase in exploration coverage throughout the entire process, highlighting the initial need for a rough understanding of the surrounding environment. 

2)During the mid-stage of exploration, our proposed method exhibit a slightly slower rate compared to FUEL. This can be attributed to our consideration of residual frontiers, which necessitates the quadrotor to make directional deviations to effectively address the uncovered areas and fill in the gaps. Additionally, all methods experience a "exploration stagnation" period due to a dead-end area in the maze map. 

3)In the final stage of exploration, our method showed a stable increase in coverage rate, However, FUEL and Rapid Frontier methods may encounter a "exploration stagnation" phase, where the coverage rate remains nearly unchanged despite an increase in distance. This occurred because these methods missed frontier in the early exploration phase, requiring them to revisit already explored areas in the final stage. These areas had fewer frontiers, resulting in a slow increase in coverage, approaching a standstill. Although the AEP method travel a shorter distance, it appear to be similar to our method, but it consume the most exploration time, as shown in Figure \ref{10}. Moreover, AEP method did not fully explore the entire space, as it still had many missed frontiers, failing to achieve complete spatial exploration.

This result validates the superiority of our proposed module, that can avoid global exploration omissions, so that exploration can steadily increase even at final stage, without revisiting the already explored area.

\subsection{Powerplant Environment Simulation}

The second simulation scenario is a powerplant scene with dimensions of $33 \times 31 \times 26 \mathrm{~m}^3$, allowing us to evaluate our Altitude-Stratified Planing strategy in a three dimensional spatial environment with significant height variations.

\begin{figure}
\begin{center}
\resizebox*{0.6\linewidth}{!}{\includegraphics{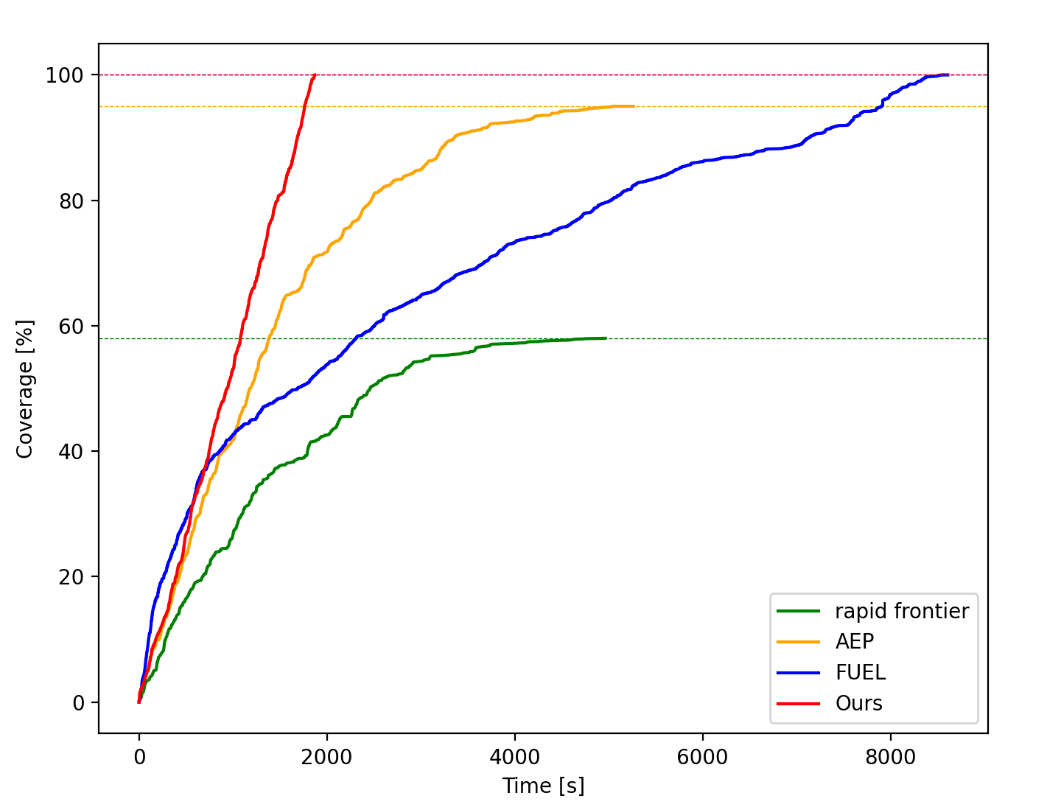}}\hspace{5pt}
\setlength{\belowcaptionskip}{-0.8cm}
\caption{Trajectories of GO-FEAP, FUEL, AEP, and Rapid Frontier in powerplant environment, showing as red, blue, green and orange respectively.}
\label{11}
\end{center}
\end{figure}

\begin{figure}
\begin{center}
\resizebox*{1.0\linewidth}{!}{\includegraphics{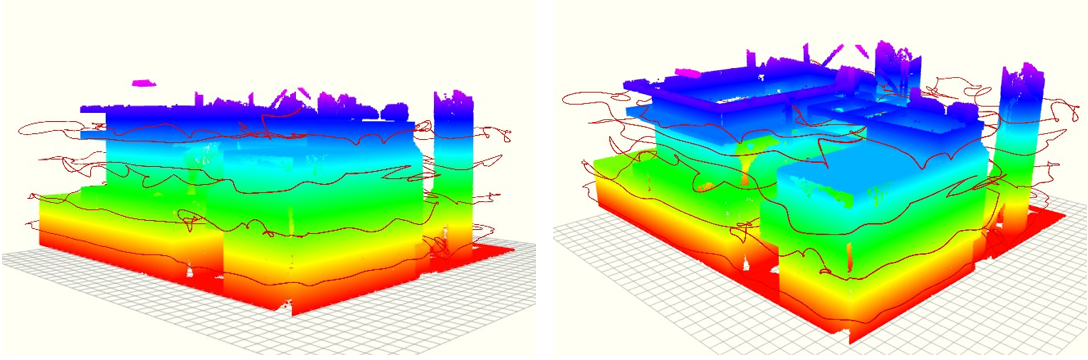}}\hspace{5pt}
\setlength{\belowcaptionskip}{-0.3cm}
\caption{Our approach's trajectory in powerplant environment.}
\label{15}
\end{center}
\end{figure}

Since the rapid method could not complete exploration when the frontier information reaches a certain threshold, we did not further compare its performance. The other results are statistically recorded in the Table \ref{tabplant}. The exploration completion progress over time is depicted in Figure \ref{11}, and the exploration trajectories are illustrated in the Figure \ref{15}.

\begin{table}[]
\caption{Exploration data statistics in powerplant environment.}
\label{tabplant}
\renewcommand{\arraystretch}{1.2}
\begin{tabular}{p{1.2cm} p{0.42cm} p{0.42cm} p{0.42cm} p{0.42cm} p{0.42cm} p{0.42cm} p{0.42cm} p{0.42cm} ccccccccc}
\hline
\multirow{2}{*}{\textbf{Method}} & \multicolumn{4}{c}{\textbf{Exploration time}}             & \multicolumn{4}{c}{\textbf{Flight distance}}              \\ \cline{2-9} 
                                 & \textbf{Avg} & \textbf{Std} & \textbf{Max} & \textbf{Min} & \textbf{Avg} & \textbf{Std} & \textbf{Max} & \textbf{Min} \\ \hline
AEP{}                        & 5697         & 568          & 6114         & 5049         & 3977         & 250          & 4245         & 3748           \\ 
FUEL{}                       & 7571         & 1830         & 8672         & 5459         & 2182         & 128          & 2321         & 2068       \\
Ours{}                       & 2167         & 95           & 2253         & 2065         & 3534         & 79           & 3621         & 3456      \\ \hline
\end{tabular}
\end{table}

\begin{table}[]
\renewcommand{\arraystretch}{1.2}
\caption{Block time comparison: GO-FEAP vs. FUEL}
\begin{tabular}{ccccccc}

\hline

\multirow{2}{*}{\textbf{Scene}} & \multirow{2}{*}{\textbf{Method}} & \multicolumn{5}{c}{\textbf{Average compute time   (ms)}} \\ \cline{3-7} 
                                &                                  & Frontier   & Global   & Local   & Traj   & Total   \\ \hline
\multirow{2}{*}{Maze}           & FUEL                             & 17               & 2        & 9       & 5      & 33      \\ 
                                & \textbf{Ours}                    & 12               & 1        & 7       & 5      & 25      \\ \hline
\multirow{2}{*}{Powerplant}     & FUEL                             & 1061             & 2425     & 410     & 5      & 3901    \\ 
                                & \textbf{Ours}                    & 184              & 9        & 11      & 5      & 209     \\ \hline
\end{tabular}
\end{table}

We calculate and compare the computation times of the two methods in Table IV, focusing primarily on block-level times. 
It can be observed that as the scene expands to a larger scale, our method demonstrates significant computation advantages, particularly in terms of frontier information computation and global planning. 
As the scene scales up, our GO-FEAP method exhibits substantial computation advantages, especially in the realms of frontier information computation and global planning. This efficiency stems from our hierarchical and altitude stratified planning strategy, which effectively constrain the upper limit on the number of frontier. Consequently, our method significantly slashes computation time for both frontier information and global planning.

In summary, 
the results from maze simulation scenarios demonstrate that GO-FEAP is feasible and effective.
the results from powerplant simulation scenarios show that the Altitude-Stratified Planning strategy effectively reduces the number of frontier that need to be maintained, enabling the handling of large-scale scenarios without significant increase in runtime. Additionally, the comparison of results between the Maze and powerplant scenarios reveals the excellent adaptability of our proposed GO-FEAP. 

\section{Real World Experiments}
To further validate the proposed method, we conduct practical indoor environment experiments. A specially customized quadrotor platform equipped with an Intel D435i depth sensor and the de next-TGU8 carrying the 11th generation Intel Core i7 processor is used. The entire system, including mapping, localization, planning, and control, is operated on the onboard computer of the quadrotor. Localization is achieved through a visual-inertial state estimator, while control trajectories are tracked using a geometric controller. 

\begin{figure}
\begin{center}
\resizebox*{1.0\linewidth}{!}{\includegraphics{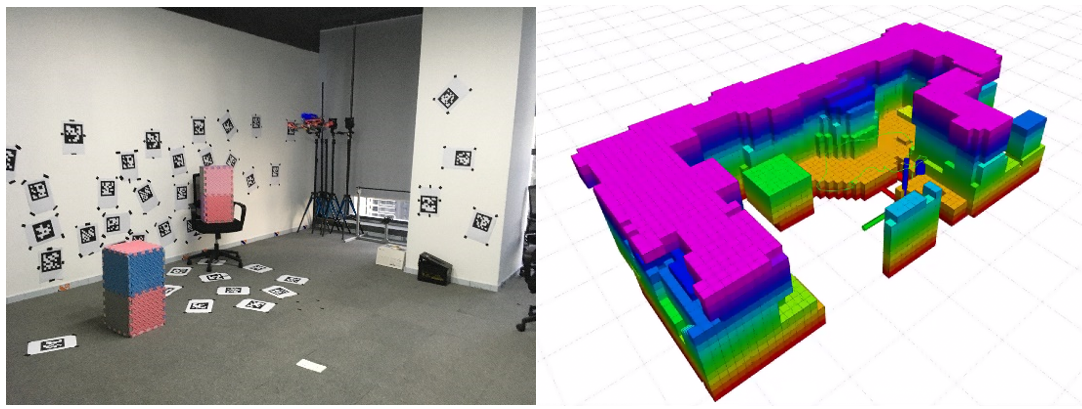}}\hspace{5pt}
\setlength{\belowcaptionskip}{-0.9cm}
\caption{Exploration experiment conducted in a indoor room.}
\label{17}
\end{center}
\end{figure}

The experiment is conducted in a room measuring $5.5 \times 5.5 \times 4 \mathrm{~m}^3$. Two obstacles are placed inside the room to create additional frontiers in space and force the quadrotor to navigate around them. The generated map is illustrated in Figure \ref{17}, and the experimental run is depicted in the accompanying figure.

\section{CONCLUSIONS}
In this study, we introduce a global optimal exploration planner that considers eliminating omission regions during exploration. We also develop an altitude-stratified planning strategy tailored for large-scale 3D environment, improving both exploration efficiency and computational cost, as demonstrated in simulation. Real-world experiments confirm the feasibility of using this strategy in practical applications, achieving real-time on-board implementation on a Unmanned Aerial Vehicle (UAV).

One limitation of our approach is the generation of significant residual frontier information when increasing the exploration layer height. While this information can be used as prior knowledge to guide exploration, it may also result in time inefficiency. Future work will address the computational challenges associated with residual frontier data and aim to further enhance exploration speed. Additional improvements in our planned future work include incorporating other sensors such as LiDAR to enhance the robustness of real-world operations and the ability to explore environment.
\clearpage
\bibliographystyle{unsrt}
\bibliography{ref}
\clearpage

\end{document}